\documentclass[conference]{IEEEtran}
\IEEEoverridecommandlockouts
\usepackage{amsmath,amssymb,amsfonts}
\usepackage{algorithmic}
\usepackage{graphicx}
\usepackage{textcomp}
\usepackage{xcolor}
\usepackage{hyperref}

\usepackage[numbers,sort&compress]{natbib}

\def\BibTeX{{\rm B\kern-.05em{\sc i\kern-.025em b}\kern-.08em
    T\kern-.1667em\lower.7ex\hbox{E}\kern-.125emX}}
\begin{document}

\title{Generation, Evaluation, and Explanation of Novelists’ Styles with Single-Token Prompts\thanks{This work builds on and substantially extends preliminary research presented at a student research workshop \citep{rezaei-2025-proposal}.}}

\author{
\IEEEauthorblockN{Mosab Rezaei, Mina Rajaei Moghadam, Abdul Rahman Shaikh, Hamed Alhoori, Reva Freedman}
\\
\IEEEauthorblockA{\textit{Department of Computer Science} \\
\textit{Northern Illinois University}\\
}
}

\maketitle

\begin{abstract}
Recent advances in large language models have created new opportunities for stylometry, the study of writing styles and authorship. Two challenges, however, remain central: training generative models when no paired data exist, and evaluating stylistic text without relying only on human judgment. In this work, we present a framework for both generating and evaluating sentences in the style of 19th-century novelists. Large language models are fine-tuned with minimal, single-token prompts to produce text in the voices of authors such as Dickens, Austen, Twain, Alcott, and Melville. To assess these generative models, we employ a transformer-based detector trained on authentic sentences, using it both as a classifier and as a tool for stylistic explanation. We complement this with syntactic comparisons and explainable AI methods, including attention-based and gradient-based analyses, to identify the linguistic cues that drive stylistic imitation. Our findings show that the generated text reflects the authors’ distinctive patterns and that AI-based evaluation offers a reliable alternative to human assessment. All artifacts of this work are published online.

\end{abstract}


\section{Introduction}

The ability to recognize and reproduce an author’s writing style has long fascinated both literary scholars and computer scientists. Stylometry, the quantitative study of writing style, rests on the idea that every author leaves behind unconscious patterns in vocabulary, syntax, and rhythm \citep{holmes1998evolution,yang2008intelligence}. These patterns have been analyzed for centuries in questions of disputed authorship, the study of literary traditions, and more recently in applications such as security and forensics \citep{wayman2009technology}. As digital text has grown in scale and diversity, stylometric methods have become increasingly valuable, offering insights not only into literature but also law and education. The expansion of online archives has broadened stylometric research, making it possible to revisit old questions with new computational tools.  

Digital libraries have played a central role in this progress. Large-scale digitization has made it possible to study thousands of historical works, while also raising questions about how to curate, analyze, and present such collections. Computational tools that detect or generate stylistic variations create opportunities for exploration, pedagogy, and access. For example, a system that generates text in the style of Dickens or Austen could enrich exhibits, classroom teaching, or reading companions that make historical literature engaging for modern audiences. More broadly, this technology shows how digital libraries are evolving from static repositories into dynamic platforms for cultural discovery and scholarly experimentation. These connections show that advances in language technologies are not only technical but also cultural, shaping how heritage is preserved and experienced.  

Recent advances in large language models shift the focus from style recognition to style generation. The core question is no longer only, \emph{Can we tell whether a passage was written by Dickens or Austen?} but also, \emph{Can a model convincingly write like them?} Style-based generation has applications in authorship verification, creative writing, adaptive education, and forensic analysis. Yet two problems remain. First, most data are unpaired, meaning there is no parallel corpus expressing the same content in different authors’ voices. Second, evaluating style is difficult. Metrics such as accuracy or BLEU capture surface resemblance but not deeper stylistic cues, while human judgments are costly and inconsistent.  

This paper proposes a framework to address these challenges. We fine-tune large language models to generate sentences in the styles of five 19th-century novelists, using single-token prompts as minimal stylistic signals. To evaluate the generated text, we train a transformer-based detector on authentic sentences, using it as both classifier and interpretive tool. We further apply explainable AI (XAI) techniques, including attention- and gradient-based analyses, to uncover the cues that drive stylistic imitation. Finally, we compare two fine-tuning approaches, full fine-tuning and LoRA, to examine trade-offs between efficiency and stylistic fidelity. The dataset\footnote{https://huggingface.co/datasets/Mosab-Rezaei/19th-century-novelists} and all artifacts\footnote{https://github.com/mosabrezaei/Text-Generation-XAI} of this work are also publicly available online.

This study is guided by two main questions:  

\vspace{1ex}  
\noindent{\textbf{RQ1:}} Given the lack of paired data, can models be trained to generate sentences that reliably reflect the styles of different authors?  

\vspace{1ex}  
\noindent{\textbf{RQ2:}} How can detector models and XAI techniques provide meaningful evaluation and explanation of stylistic generation?  

\vspace{1ex}  
In addressing these questions, our contributions are threefold. First, we propose a framework for stylistic text generation without paired data, relying on single-token prompts as minimal signals. Second, we show how transformer-based detectors combined with XAI methods can provide reliable evaluation and interpretable insights into stylistic differences. Third, we show how these methods can support analysis, preservation, and creative exploration of literary style at scale.

\section{Related Work}

Research on style transfer and stylistic text generation has followed several complementary directions. Early work focused on generative models that modify sentence style based on categorical attributes. For example, Logeswaran et al. \cite{logeswaran2018content} proposed a framework for content-preserving style transfer, while \citep{de2021formalstyler} fine-tuned GPT-2 on the GYAFC corpus to transform informal text into formal sentences. Structured approaches, such as attentional auto-encoders combined with style classifiers \cite{tian2018structured}, introduced mechanisms to maintain syntactic consistency by minimizing differences in part-of-speech structures between input and output sentences. This early stream of research established the foundation for later methods that increasingly sought to balance content fidelity with stylistic variation.  

Other studies have relied on adversarial learning. Lai et al. \cite{lai2019multiple} introduced a word-level conditional GAN with a two-phase training scheme, using dual discriminators to balance content preservation and style transfer in non-parallel text settings. Lee \cite{lee2020stable} proposed the Stable Style Transformer, a delete-and-generate framework designed to improve robustness across evaluation metrics on non-parallel datasets. Variational autoencoder extensions such as \cite{hu2017toward} explored controllable text generation by disentangling latent content and style variables. GAN-based architectures like CTERM-GAN \citep{betti-etal-2020-controlled} incorporated syntactic and semantic constraints to enhance stylistic fidelity in controlled generation. More recently, diffusion-based approaches have emerged: Diffusion-LM \citep{li2022diffusion} showed strong results in controllable text generation, and subsequent extensions such as fine-grained control via diffusion models \citep{lyu2023fine} have further improved performance in low-resource style transfer scenarios. Together, these lines of work illustrate the gradual shift from discrete, attribute-driven models toward more flexible and powerful generative frameworks.

Alongside advances in generation, efficient fine-tuning methods have gained attention.  
Low-Rank Adaptation (LoRA) \citep{hu2022lora} freezes the pre-trained model weights and injects trainable low-rank matrices into each transformer layer, reducing the number of trainable parameters while introducing no additional inference cost.  
Quantized LoRA (QLoRA) \citep{dettmers2023qlora} combines low-rank adaptation with 4-bit quantization, enabling fine-tuning of very large language models on a single GPU with minimal loss in accuracy. 
Decomposed Rank-One Adaptation (DoRA) \citep{liu2024dora} extends this line of work by separating weight updates into magnitude and direction, which improves fine-tuning stability and efficiency without adding inference overhead. These developments show how advances in parameter-efficient training make it possible to experiment with stylistic generation even under resource constraints.

Explainable AI (XAI) has become central to evaluating and interpreting modern language models. Broadly, methods can be divided into transparent approaches, which are interpretable by design, and post-hoc approaches, which explain complex neural architectures \citep{gohel2021explainable}. Post-hoc techniques include model-specific methods such as relevance scoring and model-agnostic frameworks such as LIME \citep{ribeiro2016should} and SHAP \citep{lundberg2017shap}. Comparative studies have highlighted the strengths and limitations of these techniques across domains: for example, Layer-wise Relevance Propagation (LRP) has often been found to yield more human-readable explanations than sensitivity-based methods \citep{samek2017explainable}, while recent work has systematically compared local, cohort, and global explanation strategies, underscoring the trade-offs between interpretability and fidelity \citep{cesarini2024explainable}. This strand of research emphasizes that the value of generative systems lies not only in what they produce but also in how their decisions can be explained and understood.

XAI has also been applied to text classification and generation. The Generative Explanation Framework (GEF) by Liu, Yin, and Wang \citep{liu2018towards} introduced the idea of producing fine-grained, human-readable explanations alongside predictions. More recent work combines neural architectures with explanation techniques, including hybrid classification models, LIME-based evaluations of traditional ML systems that expose gaps between accuracy and interpretability \citep{zahoor2024evaluating}, and frameworks for detecting AI-generated text that integrate stylistic features with XAI methods \citep{shah2023detecting}. Related approaches in recommender systems \citep{ouyang2018improving} similarly illustrate how explanation can enhance both transparency and predictive performance. These contributions highlight the breadth of XAI applications and their potential to bridge the gap between black-box models and human interpretability.

Research in digital libraries has long studied authorship, provenance, and integrity of textual collections. Prior work has addressed scientific authorship attribution and clustering at scale, for example, through the SMAuC corpus, which provides more than three million scientific publications with rich metadata for large-scale authorship analysis \citep{bevendorff2023smauc}. Other studies have examined plagiarism detection and paraphrase identification using neural language models, showing how transformer-based paraphrasers like BERT and RoBERTa pose new challenges for maintaining academic integrity in scholarly repositories \citep{wahle2021neural}. In addition, literature discovery and recommendation have been explored from the perspective of author preferences and stylistic features, highlighting new ways digital libraries can support navigation and engagement with scholarly texts \citep{kang2023comlittee}. Together, these threads underscore the importance of connecting authorship analysis, style modeling, and integrity tasks to the evolving needs of digital libraries.

Despite this body of work, relatively few studies explicitly connect large language models, explainable AI, and stylistic generation to digital library contexts. By situating our framework in this space, we extend the tradition of combining information retrieval, scholarly communication, and text analysis with contemporary generative and evaluative methods.

\section{Procedure}

\begin{figure*}[ht]
\centerline{\includegraphics[width=0.7\linewidth]{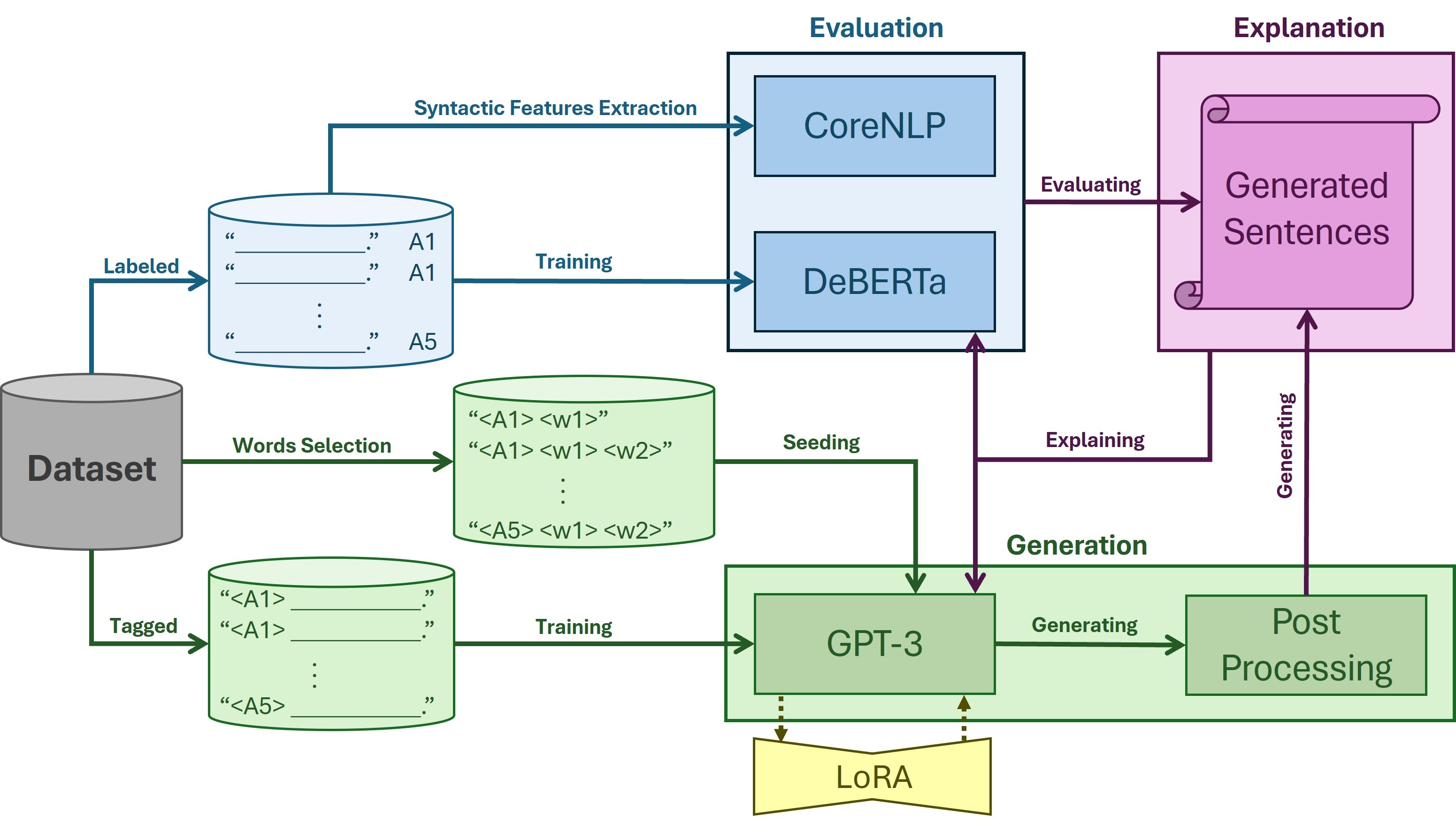}}
    \caption{The outline of the framework.}
    \label{fig:Outline}
\end{figure*}

To address our research questions, we designed a framework with two main data-flow paths: the generation path and the evaluation–explanation path (Figure \ref{fig:Outline}). The generation path (green and yellow) is responsible for producing sentences in different writing styles, while the evaluation–explanation path (blue and purple) assesses the quality of the generated outputs and explains the underlying mechanisms. The goal of the framework is to ensure that generated sentences faithfully reflect the stylistic features of target authors while also providing transparent evidence of this alignment.

\subsection{Dataset}

We constructed our dataset using texts from Project Gutenberg \citep{projectgutenberg}, focusing on five prominent 19th-century novelists: Charles Dickens, Mark Twain, Herman Melville, Jane Austen, and Louisa May Alcott. This selection balances male and female authors as well as British and American literary traditions, offering a diverse testbed for stylistic analysis.

To ensure data quality, we removed non-content pages and normalized whitespaces. Sentence segmentation was performed with the \textit{NLTK} library \citep{Bird2009NLP}, and tokenization/word counts were obtained with Stanford CoreNLP (v4.5.7) \citep{manning2014stanford}. The final dataset contains 115,471 sentences.

Unlike paired datasets (e.g., formal–informal corpora), our corpus lacks aligned sentences expressing the same content in different styles. This makes the task more challenging, as the model must learn to identify stylistic differences from sentences that differ not only in style but also in content.

To promote transparency and reproducibility, we have made the dataset publicly available as part of the paper’s artifact on GitHub. In addition to the raw sentences, the release includes rich annotations automatically extracted using Stanford CoreNLP, such as dependency relations, parse trees, and a variety of low-level and high-level syntactic features. These supplementary layers of linguistic information provide valuable resources for researchers interested not only in stylometric analysis but also in broader investigations of syntactic and semantic phenomena.

\subsection{Generation Path}

The most critical and challenging part of the framework is generating stylistically faithful text without paired samples. To address this, we introduced identifier tags at the beginning of each sentence, explicitly marking the author. These tags force the model to associate stylistic patterns with specific authors. The mapping is: \textless0\textgreater\ for Dickens, \textless1\textgreater\ for Austen, \textless2\textgreater\ for Twain, \textless3\textgreater\ for Alcott, and \textless4\textgreater\ for Melville.

We fine-tuned GPT-3–based \citep{brown2020language} models using these tags as labels. This involves using the seeding technique, which is a single token prompt approach for generating the rest of the words in a sentence. Seeds could consist of only a tag (e.g., ``\textless0\textgreater’’) or a tag followed by additional tokens (e.g., ``\textless0\textgreater\ hello’’ or ``\textless0\textgreater\ today is’’). In other words, adding additional tokens to the author tag is similar to providing a more detailed prompt. As we add more tokens, the model completes the rest of the sentence not only based on the author tag but also by strongly conditioning on the initial tokens. 

The GPT-3-based model we employed for the generation is GPT-Neo 1.3B \citep{gpt-neo}, an autoregressive model with 1.3 billion parameters developed by EleutherAI. We selected GPT-Neo because it is one of the largest publicly available open-source autoregressive models that we could reliably fine-tune and deploy within our computational resource limitations. We ran our experiments on an NVIDIA A100 GPU in Google Colab Pro \citep{googlecolab2025}, with 83.5 GB of system RAM and 40 GB of GPU RAM. 

We applied two fine-tuning methods: full fine-tuning (FFT) and Low-Rank Adaptation (LoRA) \citep{hu2022lora}. LoRA was chosen because it reduces computational and storage costs, making it an efficient alternative to FFT. Using both approaches allows us to compare trade-offs in performance and efficiency.

After fine-tuning, we generated 50,000 sentences: 25,000 with FFT and 25,000 with LoRA, corresponding to 5,000 sentences per author per method. The generator hyperparameters were set to \textit{do\_sample} = true, \textit{temperature} = 0.9, and \textit{max\_new\_tokens} = 64. Post-processing removed author tags and eliminated repeated tokens after the end tags or incomplete sentences. The result is a collection of high-quality generated sentences for downstream evaluation.

\subsection{Evaluation-Explanation Path}

Although a variety of metrics are commonly used to evaluate generator performance, including accuracy, F1 score, BLEU, and BERTScore, achieving high values on these metrics does not necessarily reflect true effectiveness in generating sentences with distinct writing styles. Moreover, these scores do not always capture the nuances of model performance. Human evaluation, while valuable, introduces additional complexity and subjectivity.

To address these challenges, we adopt a complementary evaluation pathway that combines an AI-based evaluation approach with explainable AI (XAI) techniques. First, we train a DeBERTa V3-Large classifier \citep{he2020deberta, DeBERTaV3} on sentences from the dataset, using their author labels as supervision. A highly accurate classifier serves two purposes: First, it demonstrates that authorial features are indeed distinguishable. Second, it provides a reliable evaluator of generated sentences by verifying whether they align with the intended writing style. As illustrated in the blue section of Figure \ref{fig:Outline}, this pathway also incorporates syntactic feature extraction, including both low-level \citep{moghadam2024_1} and high-level \citep{moghadam2024_2} features. These syntactic measures enable us to highlight not only the stylistic differences among authors but also the distinctions between real and generated sentences.

The purple section of Figure \ref{fig:Outline} depicts the role of XAI techniques, which are employed to provide deeper insight into these differences and to furnish evidence that the models genuinely capture patterns underlying authorial style. To test whether the model relies on author tags during generation, we employ both Attention Enrichment (AE) \citep{clark2019does, vig2019visualizing, rogers2020primer} and Integrated Gradients (IG) \citep{sundararajan2017axiomatic, kokhlikyan2020captum}.

For AE, we examine whether the author tags attract sufficient attention across the model’s layers. Specifically, we compute the “to-tag” attention mass as the average attention weight assigned to the tag span by all subsequent tokens at layer $\ell$. To account for varying tag lengths, this value is normalized by a baseline expectation (tag length divided by total sequence length), yielding the following enrichment score:

\begin{equation}
\text{Enrichment}^{(\ell)} = 
\frac{\text{to-tag mass}^{(\ell)}}{\text{tag\_len}/T}
\end{equation}

where $T$ denotes the number of valid tokens in the sequence and $tag\_len$ is the number of tag tokens. An enrichment score greater than 1.0 indicates that the tag receives disproportionately high attention compared to chance. We report only ``to-tag'' enrichment, as the ``from-tag'' metric is not interpretable in this setup; the tag is positioned at the beginning of the sequence, and forward attention from the tag to later tokens is masked.

The IG analysis provides complementary gradient-based evidence that author tags are not ignored after the initial prefix. Instead, they directly influence the probability distribution over subsequent tokens. AE shows that tags are repeatedly referenced during generation, while IG confirms that they exert a measurable influence on output selection. Given an input embedding vector $\mathbf{e}$ (prompt + generated tokens) and a baseline embedding $\mathbf{e}'$ (here, zeroed embeddings for the prompt), the attribution for dimension $\mathbf{i}'$ is defined as:

\begin{equation}
\label{eq:ig-dim}
\mathrm{IG}_i
= \big(\mathbf{e}_i - \mathbf{e'}_i\big)\,
\int_{0}^{1}
\frac{\partial f\!\big(\mathbf{e}' + \alpha(\mathbf{e}-\mathbf{e}')\big)}{\partial \mathbf{e}_i}
\, d\alpha
\end{equation}

Attributions are aggregated at the token level by taking the L2 norm across embedding dimensions:

\begin{equation}
\label{eq:token-agg}
A_{i,j} \;=\; \big\| \mathrm{IG}_{ij} \big\|_2
\end{equation}

where $A_{i,j}$ is the importance of prompt token $ti$ for the probability of generating token 
$j$.

Beyond generation analysis, we also apply token-level IG to the DeBERTa classifier in order to quantify how much each input token contributes to the model’s confidence in a given class. By aggregating token attributions across many sentences, this analysis reveals author-specific lexical signals that are useful for downstream evaluation.

\section{Results}

This section reports the outcomes of generation, evaluation, and explanation. The generation results cover the performance of both fine-tuning methods and include a human expert analysis of the generated sentences. The evaluation results examine how effectively the generator models can reproduce sentences in the style of the target authors, as assessed by the classifier model. Finally, the explanation results provide insights into the stylistic differences among authors and highlight the influence of single-token prompts on the generated outputs.

\subsection{Generation}

As described in the Procedure section, we experimented with two fine-tuning strategies: GPT-Neo with full fine-tuning (FFT) and GPT-Neo with Low-Rank Adaptation (LoRA). Table \ref{table: fine_tuning_methods} compares the computational requirements of these approaches. FFT required updating all 1.4 billion parameters, consuming approximately 44 GB of storage and 5 hours and 21 minutes of training time for three epochs. By contrast, LoRA reduced the number of trainable parameters to about 3.1 million, lowering the storage requirement to 12 GB and shortening training time to 4 hours and 52 minutes for the same number of epochs. This contrast illustrates the substantial efficiency benefits of LoRA while maintaining practical training feasibility.

\begin{table*}[ht]
\centering
\caption{Two fine-tuning methods have been used: Full Fine Tuning (FFT) and Low-Rank Adaptation (LoRA).}
\label{table: fine_tuning_methods}
\begin{tabular}{lrccc}
\hline
\multicolumn{1}{c}{Model} & \multicolumn{1}{c}{Trainable Parameters} & \multicolumn{1}{c}{Epoch} & \multicolumn{1}{c}{Training Time} & \multicolumn{1}{c}{Used Space} \\ \hline
 GPT-Neo (FFT)        & 1,418,506,240        & 3           & 5:21             & $\sim$44 GB   \\
 GPT-Neo (LoRA)       & 3,145,728            & 3           & 4:52             & $\sim$12 GB   \\ \hline
\end{tabular}
\end{table*}

Following fine-tuning, we generated 25,000 sentences with each model. Representative examples of FFT outputs are shown in Table \ref{table:generated_sentences}. A key observation is that LoRA occasionally produced excessive word repetition after the end tag identifier (\textless end\textgreater). These repetitions, however, were systematically removed during post-processing to ensure clean sentence outputs.

Beyond quantitative differences, non-systematic human expert evaluation confirmed that the fine-tuned models successfully captured author-specific stylistic features. For example, in the first sentence attributed to Charles Dickens, we observe British English usage such as ``had got'', since American English typically uses ``had gotten.'' In the second sentence attributed to Jane Austen, the reference to parties and social behavior clearly aligns with themes frequently explored in her stories. Regarding the fourth sentence by Alcott, words like ``courage'' and ``resolution'' reflect the language commonly found in novels from her period. The fifth sentence, attributed to Melville, focuses on men and boys, a theme prevalent throughout his works. In the seventh sentence attributed to Austen, it is not surprising to encounter a depiction of a woman busy shopping in the street, a typical scenario in Austen's novels. For the eighth sentence attributed to Mark Twain, the importance of boys and references to Mississippi strongly reflect his characteristic themes. The last sentence attributed to Alcott resembles a direct note to the reader, a common stylistic feature in 19th-century literature.

\begin{table*}
\centering
\caption{Several examples of generated sentences using the seeding technique for the expected writing styles.}
\label{table:generated_sentences}
\begin{tabular}{lll}
\hline
\multicolumn{1}{c}{Target} & \multicolumn{1}{c}{Seed}      & \multicolumn{1}{c}{Generated Sentences}                                   \\   \hline\hline
Charles Dickens            & \textless0\textgreater \ When & \textbf{When} I had got my breath, I said, “I am going to London.         \\
Jane Austen                & \textless1\textgreater \ When & \textbf{When} they were gone, she sat down again, ...                     \\ 
Mark Twain                 & \textless2\textgreater \ When & \textbf{When} the sun went down, we had a grand supper, ...               \\ 
Louisa May Alcott          & \textless3\textgreater \ He   & \textbf{He} was a man of great courage, and a man of great resolution.    \\
Herman Melville            & \textless4\textgreater \ He   & \textbf{He} had been a very good-looking young man, ...                   \\
Charles Dickens            & \textless0\textgreater \ He   & \textbf{He} was a man of great talent, and his music was considered ...   \\
Jane Austen                & \textless1\textgreater \ A    & \textbf{A} very few minutes more, however, and she was in the street, ... \\
Mark Twain                 & \textless2\textgreater \ A    & \textbf{A} few of the boys had gone to the river, ...                     \\
Louisa May Alcott          & \textless3\textgreater \ A    & \textbf{A} few words of explanation will make it clear.                   \\   \hline                                  
\end{tabular}
\end{table*}

\subsection{Evaluation}

In order to obtain a reliable AI-based evaluator, we fine-tuned the DeBERTa V3-Large classifier on real sentences. After nine epochs, both accuracy and F1 score stabilized at 87\% on the test set. This trained classifier was then employed to evaluate generated sentences. Each generated sentence was produced from a single-token prompt indicating the target author, followed by one or two random words. The expectation was that the classifier should assign the generated sentence to the class corresponding to the prompt tag.

\begin{figure*}[ht]    \centerline{\includegraphics[width=1\linewidth]{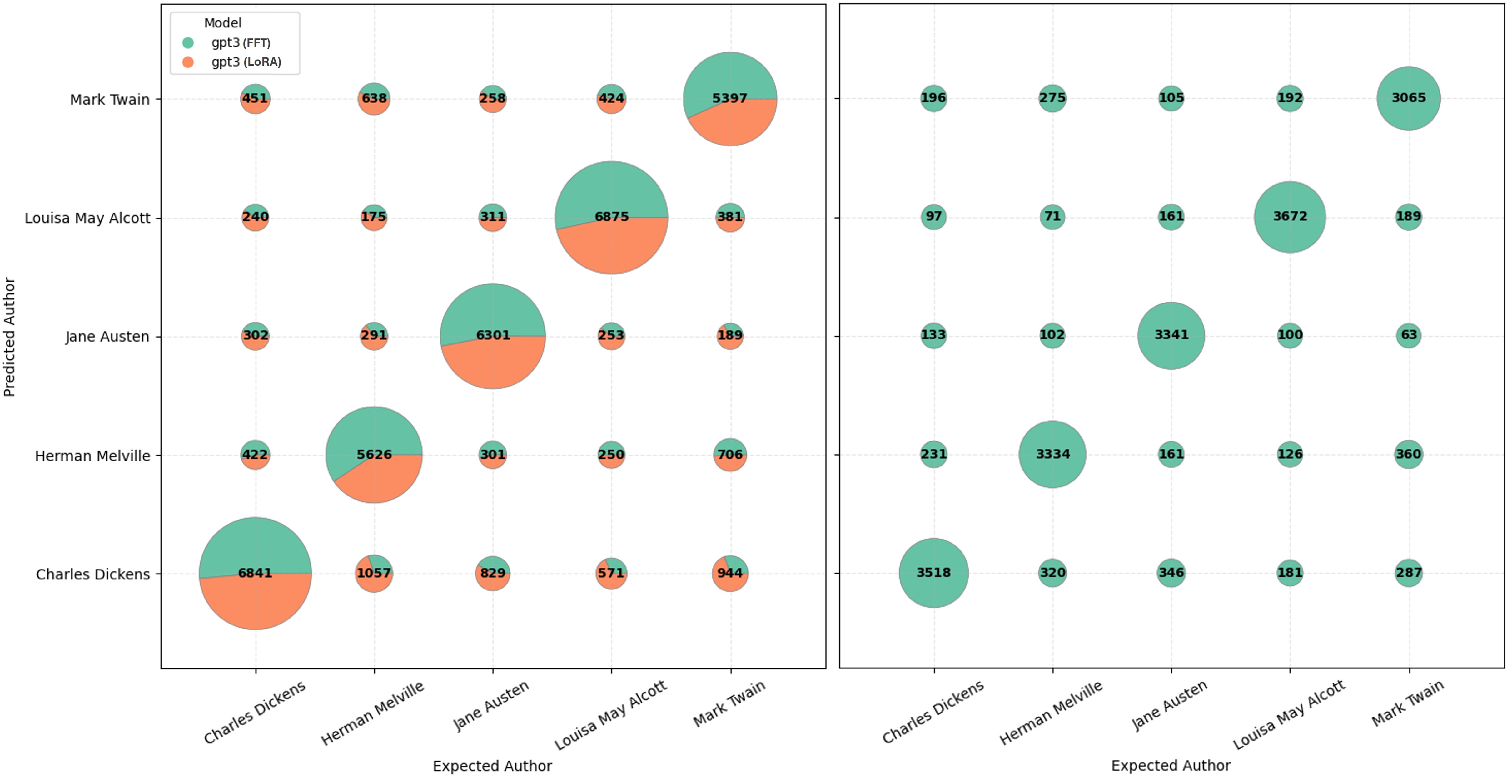}}
    \caption{Evaluating the writing style of generated sentences. Left: Both generator models, FFT and LoRA. Right: Only the FFT results}
    \label{fig:Circles}
\end{figure*}

The left panel of Figure \ref{fig:Circles} illustrates the evaluation results for two generator models: GPT-Neo fine-tuned with full fine-tuning (FFT), shown in green, and GPT-Neo fine-tuned with LoRA, shown in orange. The DeBERTa classifier aligns with the expected writing style for most sentences produced by both models. For example, among 10,000 sentences prompted for Dickens, DeBERTa (y-axis) identified 6,841 as Dickens, while the remainder were attributed to other authors. The chart demonstrates that both fine-tuning methods enable the generator to capture authorial style, but with different levels of success. GPT-Neo with FFT clearly outperforms LoRA, producing more samples along the ascending diagonal, indicating agreement between expected and predicted style, and fewer off-diagonal misclassifications. Notably, LoRA shows weaknesses in replicating the styles of Mark Twain and Herman Melville. Nevertheless, given LoRA’s significant efficiency in terms of storage requirements, it remains an attractive alternative when computational constraints are critical. Because FFT exhibited stronger overall performance, its results are reported separately in the right panel of Figure \ref{fig:Circles}.
    
Although DeBERTa was chosen for its strong classification performance, it is important to acknowledge potential sources of error. As shown in Table \ref{table: confidency}, a more reliable evaluation can be obtained by focusing on high-confidence samples, those for which DeBERTa assigned predictions with more than 93\% confidence. For these samples, the average confidence exceeded 99\%. On real data, DeBERTa achieved 93\% accuracy on 88\% of the samples, making it a robust evaluator for this study. Based on this criterion, we re-evaluated both generators. For the FFT model, DeBERTa agreed with 82\% of the labels on 81\% of the high-confidence samples. These results indicate that the generators learned to capture meaningful stylistic patterns and to mimic each author’s style with substantial accuracy, even when conditioned only on single-token prompts.

\begin{table*}[ht]
\centering
\caption{Average accuracy of high-confidence predictions by the DeBERTa V3-Large classifier.}
\label{table: confidency}
\begin{tabular}{lllll}
\hline
\multicolumn{1}{c}{DeBERTa Evaluation Mode} & \multicolumn{1}{c}{Labels} & \multicolumn{1}{c}{Number of Samples} & \multicolumn{1}{c}{Ave of Conf} & \multicolumn{1}{c}{Ave of Acc} \\ \hline
Testset Sentences                     & Groundtruth                             & 20,339 / 23,095 (88\%)                              & 99\%             & 93\%\\
Generated by GPT-Neo (FFT)               & Expected Style                                & 20,151 / 25,000 (81\%)                              & 99\%             & 82\% \\
Generated by GPT-Neo (LoRA)              & Expected Style                                & 18,693 / 25,000 (75\%)                             & 99\%             & 73\% \\ \hline
\end{tabular}
\end{table*}

\begin{figure}[ht]    \centerline{\includegraphics[width=1\linewidth]{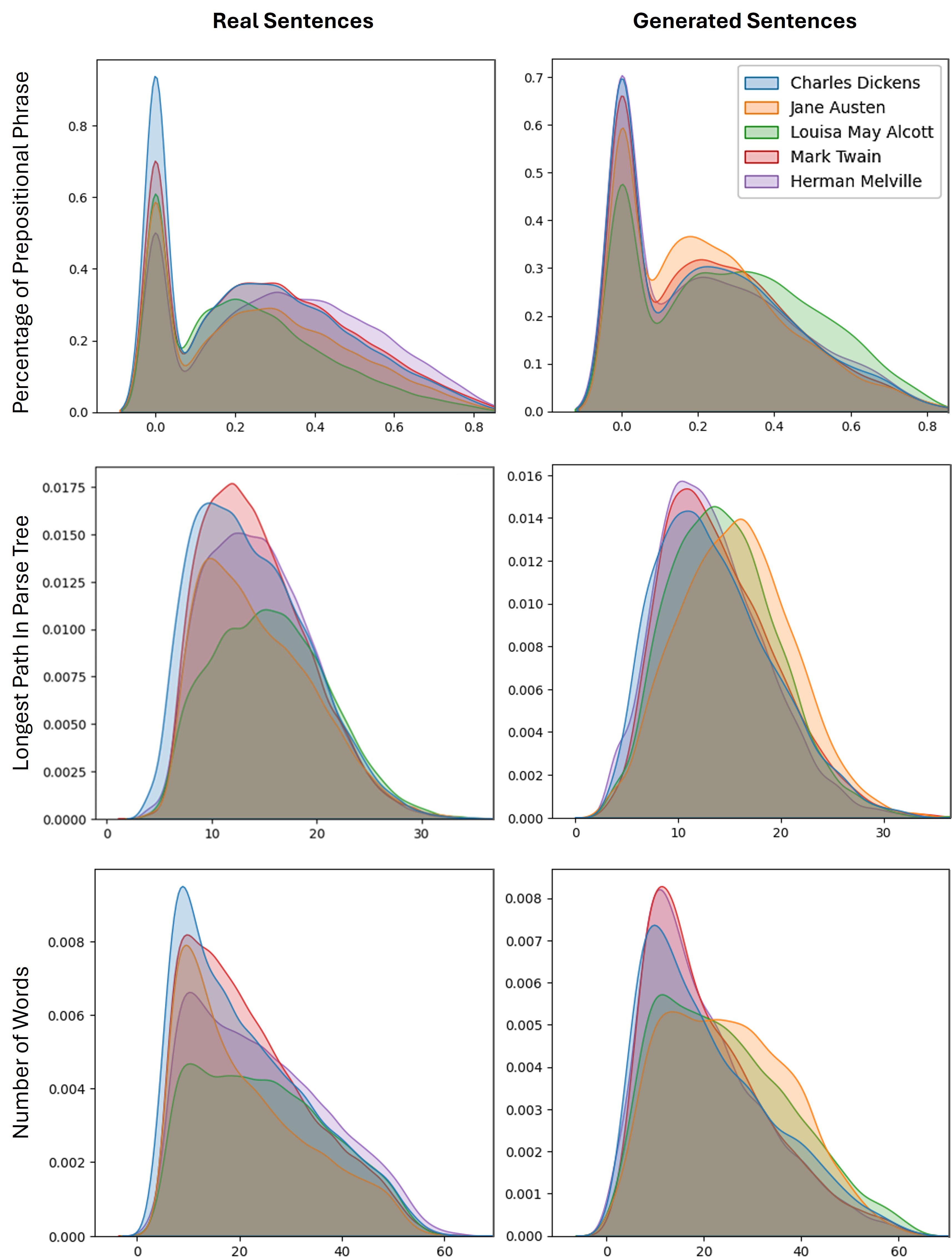}}
    \caption{Histogram of the longest path in parse trees, percentage of prepositional phrases, and number of words in real and generated sentences.}
    \label{fig:Authors_hist}
\end{figure}

Beyond classification-based evaluation, we also examined syntactic characteristics to assess how closely generated sentences by GPT-Neo FFT replicate the distributional patterns of real sentences. Figure \ref{fig:Authors_hist} presents histograms of three representative features: the percentage of prepositional phrases (first row), the longest path in the parse tree (second row), and the number of words per sentence (third row), comparing real and generated data. Differences between authors are evident. For example, in the real data, Louisa May Alcott (green) displays a distribution of parse-tree path lengths that differs markedly from Mark Twain (red). Twain’s sentences tend to have shorter paths, suggesting simpler syntactic structures, while Alcott’s sentences exhibit a more uniform spread across longer path lengths, reflecting greater syntactic complexity.

To compare real and generated data, 16 features were analyzed, three of which are presented in Figure \ref{fig:Authors_hist}. Although some differences exist, for example, the longest path in the parse tree for Alcott or the number of words per sentence for Austen, the histograms of real and generated data overall exhibit substantial similarity.

\subsection{Explanation}

AE analysis results across author tags (Dickens: \textless0\textgreater, Austen: \textless1\textgreater, Twain: \textless2\textgreater) for GPT-Neo FFT are reported in Table \ref{table: layer_attentions}. We observed consistent enrichment in the middle and higher transformer layers. While the earliest layers remained close to baseline, layers around 10 and 22 exhibited pronounced amplification. For instance, for the \textless0\textgreater tag, the ``to\_tag'' attention rose from 0.04 in layer 0 to a maximum of 0.26 in layer 20. The enrichment factor also increased steadily, reaching up to 4.4× the baseline in layers 20 and 22. Although the first layers devoted very little attention to the tags, the deeper layers consistently allocated between 10\% and 28\% of their attention mass, far exceeding the 3\% expected by random chance. This pattern suggests that, during text generation, the model repeatedly re-references with the author tag in its deeper layers, thereby reinforcing the intended stylistic conditioning.

\begin{table}[ht]
\centering
\caption{Attention Enrichment (AE): Quantified self-attention to the author tag (e.g., ``\textless 0\textgreater'', ``\textless 1\textgreater'', and ``\textless 2\textgreater'') across all transformer layers during generation. The reported numbers are ``to\_tag'' values, and the numbers in parentheses are Enrichment values.}
\label{table: layer_attentions}
\begin{tabular}{llll}
\hline
\multicolumn{1}{c}{Layer} & \multicolumn{1}{c}{\textless 0\textgreater} & \multicolumn{1}{c}{\textless 1\textgreater} & \multicolumn{1}{c}{\textless 2\textgreater} \\ \hline
00       & 0.04 (x0.6)        & 0.05 (x0.7)       & 0.03 (x0.6)\\
01       & 0.09 (x1.5)        & 0.10 (x1.5)       & 0.07 (x1.2)\\
02       & 0.12 (x2.2)        & 0.14 (x2.0)       & 0.09 (x1.5)\\
03       & 0.09 (x1.5)        & 0.09 (x1.3)       & 0.06 (x1.0)\\
04       & 0.05 (x1.5)        & 0.05 (x0.8)       & 0.03 (x0.6)\\
05       & 0.04 (x0.7)        & 0.05 (x0.7)       & 0.03 (x0.5)\\
06       & 0.05 (x0.9)        & 0.06 (x0.9)       & 0.03 (x0.6)\\
07       & 0.04 (x0.8)        & 0.05 (x0.7)       & 0.03 (x0.5)\\
08       & 0.12 (x2.2)        & 0.13 (x1.9)       & 0.05 (x1.0)\\
09       & 0.07 (x1.2)        & 0.06 (x0.9)       & 0.03 (x0.6)\\
10       & 0.23 (x4.2)        & 0.23 (x3.3)       & 0.12 (x2.1)\\
11       & 0.11 (x2.0)        & 0.09 (x1.3)       & 0.05 (x1.0)\\
12       & 0.22 (x3.9)        & 0.23 (x3.4)       & 0.14 (x2.4)\\
13       & 0.11 (x2.0)        & 0.09 (x1.3)       & 0.06 (x1.0)\\
14       & 0.24 (x4.2)        & 0.28 (x4.0)       & 0.16 (x2.7)\\
15       & 0.11 (x2.0)        & 0.13 (x1.9)       & 0.06 (x1.1)\\
16       & 0.24 (x4.2)        & 0.29 (x4.2)       & 0.13 (x2.3)\\
17       & 0.13 (x2.3)        & 0.16 (x2.3)       & 0.07 (x1.2)\\
18       & 0.23 (x4.1)        & 0.29 (x4.2)       & 0.15 (x2.6)\\
19       & 0.16 (x2.8)        & 0.21 (x3.1)       & 0.08 (x1.4)\\
20       & 0.26 (x4.6)        & 0.29 (x4.2)       & 0.17 (x2.9)\\
21       & 0.16 (x2.8)        & 0.20 (x2.9)       & 0.09 (x1.6)\\
22       & 0.25 (x4.4)        & 0.28 (x4.0)       & 0.16 (x2.8)\\
23       & 0.10 (x1.9)        & 0.11 (x1.6)       & 0.06 (x1.1)\\\hline
\end{tabular}
\end{table}

\begin{figure}[ht]    \centerline{\includegraphics[width=1\linewidth]{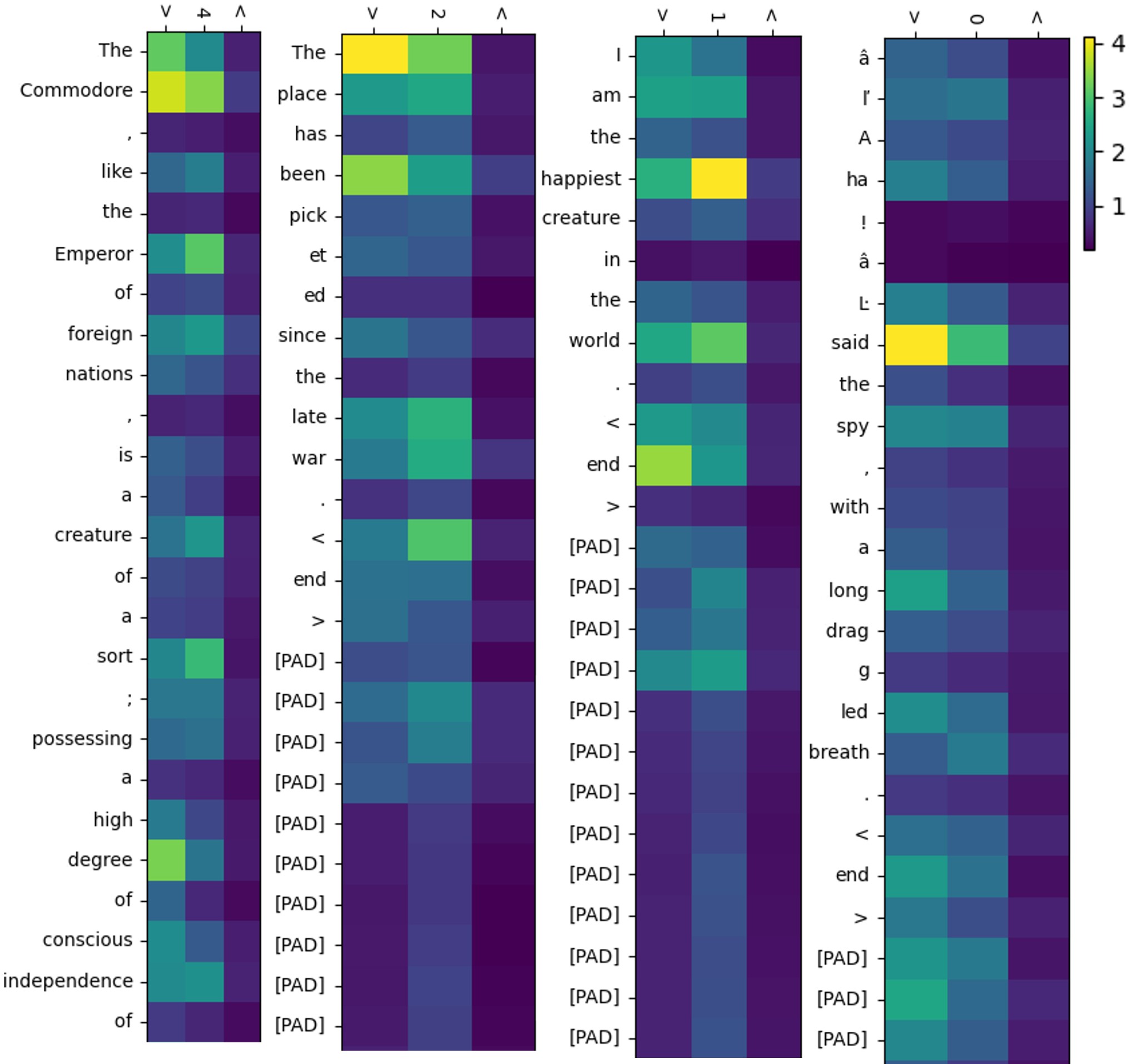}}
    \caption{Integrated Gradients (IG): Five examples of the tag identifier effect on the next token prediction. Generated sentences are:\\
    \textless 0\textgreater ``Aha!'' said the spy, with a long draggled breath.\\
    \textless 1\textgreater I am the happiest creature in the world.\\
    \textless 2\textgreater The place has been picketed since the late war.\\
    \textless 4\textgreater The Commodore, like the Emperor of foreign nations, is a ...}
    \label{fig:tags_effects}
\end{figure}

Figure \ref{fig:tags_effects} also presents IG heatmaps for each author tag for GPT-Neo FFT. In these plots, columns correspond to prompt tokens (the tags) and rows correspond to generated tokens. Brighter regions indicate stronger attribution magnitudes, meaning that the associated prompt token contributed more strongly to the probability of generating the corresponding output token. Because the symbol ``\textless'' always appears as the first token at the beginning of every sentence, it exerts almost no influence on generation. By contrast, the other tokens (``0\textgreater'', ``1\textgreater'', ``2\textgreater'', ``3\textgreater'', and ``4\textgreater'') exhibit a significant impact on output probabilities. In particular, the consistently high attribution magnitudes for these tags across all five authors highlight that the model is using them as effective control signals, ensuring that stylistic conditioning is actively enforced throughout the generation process.

\begin{figure*}[ht]    \centerline{\includegraphics[width=1\linewidth]{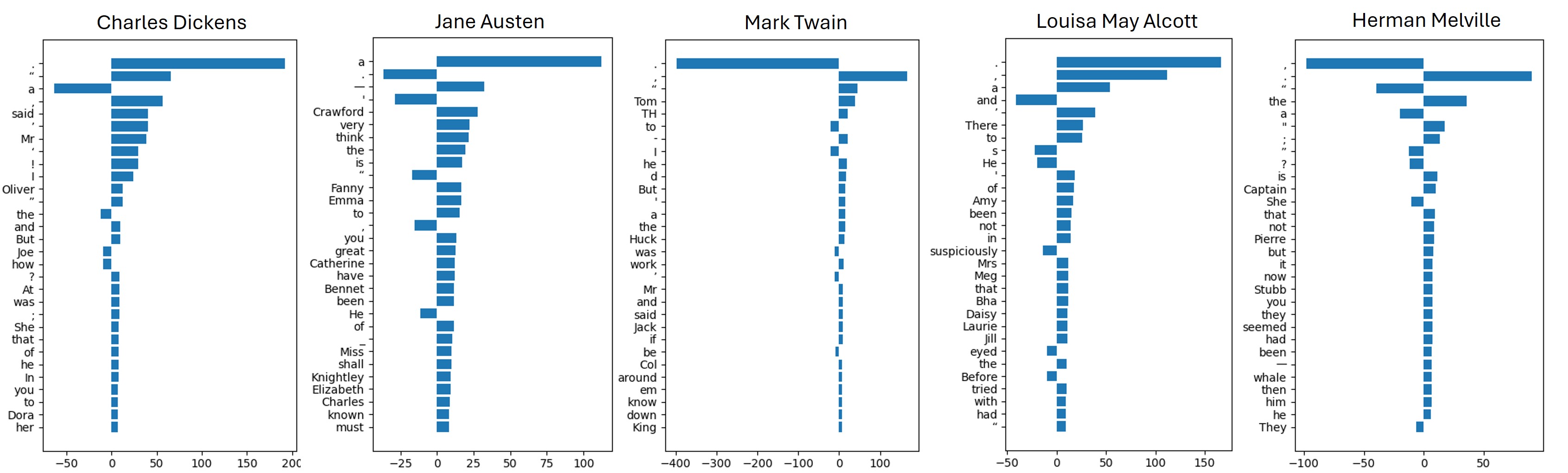}}
    \caption{Token-Level Attributions with Integrated Gradients.}
    \label{fig:Top_tokens}
\end{figure*}

The token-level analysis of the DeBERTa classifier highlights which tokens most strongly influenced the classifier’s decision for each author. The results in Figure \ref{fig:Top_tokens} demonstrate clear author-specific signals, which are explained below for each author:

The attributions for Charles Dickens are dominated by punctuation and dialogue markers such as quotation marks, commas, ``said'', and ``Mr'', alongside character names including ``Oliver'', ``Joe'', and ``Dora''. This pattern reflects Dickens’s narrative style, which is heavily dialogue-driven and frequently introduces characters by name. The model clearly identifies his stylistic reliance on conversations and character interactions as key signals.

The strongest attributions for Jane Austen are linked to social and familial names such as ``Crawford'', ``Knightley'', ``Catherine'', ``Bennet'', and ``Elizabeth'', as well as verbs like ``think'' and ``shall''. These tokens capture her focus on family relations, social hierarchy, and interpersonal dynamics. Therefore, the classifier recognizes Austen’s style through proper nouns and conversational vocabulary tied to her themes of manners and society.

Mark Twain’s attributions highlight character names like ``Tom'', ``Huck'', ``King'', and ``Jack'', along with dialectal tokens such as ``em'' and ``Col''. In contrast, punctuation and generic function words often receive negative contributions, showing that the model downplays syntax in favor of thematic content. These results align with Twain’s distinctive style, which blends colloquial dialogue and vivid character voices to establish his narrative identity.

Louisa May Alcott’s most influential tokens are family and domestic names such as ``Amy'', ``Meg'', ``Laurie'', ``Daisy'', and ``Jill'', complemented by relational markers like ``Mrs'' and dialogue connectors such as ``and'' and ``He''. The prominence of these terms reflects her focus on family life, relationships, and moral development, especially in Little Women. The classifier identifies her style primarily through this vocabulary of kinship and household dynamics.

Herman Melville’s results emphasize thematic nouns and names, most notably ``Captain'', ``Pierre'', ``Stubbs'', and ``whale'', alongside general function words like ``the'' and ``a''. The strong attribution of marine tokens reflects the maritime setting of works such as Moby-Dick, while the presence of both positive and negative contributions for common words indicates that the model balances thematic content with baseline syntactic structure. His style is therefore recognized through a combination of seafaring vocabulary and distinctive character references.

\section{Limitation and Future Work}

While our evaluation relied primarily on a single classifier (DeBERTa V3-Large), future work can broaden evaluation in several ways. First, adding multiple evaluators would reduce bias from relying on one classifier. Second, we plan to experiment with zero-shot judges, where an instruction-tuned LLM is directly asked to identify or rate the stylistic author of a generated sentence (e.g., “Does this sentence sound like Dickens, Austen, Twain, Alcott, or Melville?”). Third, incorporating a small human study with blind annotation can provide a complementary perspective and strengthen reliability.

Although the central aim of this work is to propose a generalizable framework for style generation with unpaired samples, our current experiments are limited to a dataset containing only five novelists. In future work, we plan to expand this evaluation to more diverse corpora, including scholarly datasets with hundreds of authors. Such an expansion would allow for a richer assessment of stylistic diversity and provide deeper insights into the adaptability of the proposed framework.

Another limitation observed in the results is that the evaluator model often placed considerable emphasis on specific character names frequently used by authors in their novels. While such lexical cues can be considered part of an author’s stylistic signature, this reliance risks overshadowing more substantive features of style, such as syntax, rhythm, or discourse structure. To mitigate this, we plan to update the dataset to reduce name dependency, thereby encouraging the model to capture deeper and more complex stylistic patterns that more faithfully reflect authorial style.

\section{Conclusion}

This study demonstrates that large language models can be effectively fine-tuned to generate text in the distinct styles of 19th-century novelists using minimal prompting. We compared two fine-tuning approaches, Full Fine-Tuning (FFT) and Low-Rank Adaptation (LoRA), and found that although LoRA offers substantial efficiency gains by reducing the number of trainable parameters, FFT achieves greater stylistic fidelity. Human evaluation further confirmed that the generated sentences captured thematic and linguistic patterns consistent with the targeted authors.

To quantify stylistic accuracy, we employed a fine-tuned DeBERTa V3-Large classifier as an evaluator. Results showed that generated outputs were classified as the intended author in the majority of cases, with FFT outperforming LoRA, particularly for more challenging authors such as Twain and Melville. When restricting evaluation to high-confidence predictions, FFT achieved 82\% agreement between generator prompts and classifier decisions, further validating its ability to capture authorial patterns.

The comparison of syntactic features between real and generated data revealed that, despite some author-specific deviations, such as longer parse tree paths for Alcott and a reduced proportion of short sentences for Austen, the generators largely reproduced the overall distributional patterns of both low-level and high-level syntactic features. The close alignment of histograms across real and generated texts indicates that the models are capable of capturing and mimicking the syntactic structure underlying each author’s writing style.

Our interpretability analyses provide additional evidence of this effect. Attention Enrichment revealed that deeper transformer layers consistently re-referenced author tags, with up to 28\% of the attention mass devoted to these tokens, confirming their role as reliable conditioning signals. Integrated Gradients analyses supported this observation, showing that tag identifiers carried substantial attribution magnitudes across generations.

Finally, token-level attribution with the DeBERTa classifier highlighted distinctive lexical markers characteristic of each author: Dickens’s reliance on dialogue and character names, Austen’s social and familial references, Twain’s protagonist names and dialectal expressions, Alcott’s family-centered vocabulary, and Melville’s maritime themes. These results indicate that the generators not only replicate broad stylistic tendencies but also reproduce fine-grained lexical cues associated with individual authors.

In sum, our findings demonstrate that even with single-token prompts, GPT-3-based models fine-tuned on literary data can generate sentences that convincingly reflect the writing styles of 19th-century novelists, with FFT providing the most faithful stylistic reproduction. By integrating human evaluation with AI-based classification and interpretability techniques such as Attention Enrichment and Integrated Gradients, this study offers both empirical evidence and explanatory insights into how modern generative models capture and reproduce literary style.

\bibliographystyle{IEEEtran}
\bibliography{my_bibliography}

\end{document}